\documentclass[11pt,a4paper]{article}
\usepackage[hyperref]{naaclhlt2018}
\usepackage{times}
\usepackage{latexsym}
\usepackage{amsmath}
\usepackage{amssymb}
\usepackage{color,colortbl}
\usepackage{booktabs}
\usepackage{subcaption}
\usepackage{multirow}
\usepackage{url}
\usepackage{ifthen}
\usepackage{xspace}
\usepackage{graphicx}
\usepackage{framed}
\usepackage{textcomp}
\usepackage{float}
\aclfinalcopy % Uncomment this line for the final submission
\newif\ifcomment
\commentfalse

\newcommand\p[1]{\ensuremath{\left( #1 \right)}} % Parenthesis ()
\newcommand\refsec[1]{Section~\ref{sec:#1}}

\newcommand\reffig[1]{Figure~\ref{fig:#1}}

\newcommand\reftab[1]{Table~\ref{tab:#1}}

\ifthenelse{\isundefined{\definition}}{\newtheorem{definition}{Definition}}{}
\ifthenelse{\isundefined{\assumption}}{\newtheorem{assumption}{Assumption}}{}
\ifthenelse{\isundefined{\hypothesis}}{\newtheorem{hypothesis}{Hypothesis}}{}
\ifthenelse{\isundefined{\proposition}}{\newtheorem{proposition}{Proposition}}{}
\ifthenelse{\isundefined{\theorem}}{\newtheorem{theorem}{Theorem}}{}
\ifthenelse{\isundefined{\lemma}}{\newtheorem{lemma}{Lemma}}{}
\ifthenelse{\isundefined{\corollary}}{\newtheorem{corollary}{Corollary}}{}
\ifthenelse{\isundefined{\alg}}{\newtheorem{alg}{Algorithm}}{}
\ifthenelse{\isundefined{\example}}{\newtheorem{example}{Example}}{}
\definecolor{darkgreen}{rgb}{0,0.5,0}
\ifcomment
\newcommand\pl[1]{\textcolor{red}{[PL: #1]}}
\newcommand\hh[1]{\textcolor{blue}{[HH: #1]}}
\newcommand\rj[1]{\textcolor{darkgreen}{[RJ: #1]}}
\newcommand\jcl[1]{\textcolor{orange}{[JCL: #1]}}
\else
\newcommand\pl[1]{}
\newcommand\hh[1]{}
\newcommand\rj[1]{}
\newcommand\jcl[1]{}

\fi

\newcommand{\yelp}{\textsc{Yelp}\xspace}
\newcommand{\amazon}{\textsc{Amazon}\xspace}
\newcommand{\captions}{\textsc{Captions}\xspace}
\newcommand{\source}{\textsc{Source}\xspace}
\newcommand{\crossalign}{\textsc{CrossAligned}\xspace}
\newcommand{\fader}{\textsc{StyleEmbedding}\xspace}
\newcommand{\multidecoder}{\textsc{MultiDecoder}\xspace}
\newcommand{\rulebased}{\textsc{TemplateBased}\xspace}
\newcommand{\retrieval}{\textsc{RetrieveOnly}\xspace}
\newcommand{\original}{\textsc{DeleteAndRetrieve}\xspace}
\newcommand{\ylabel}{\textsc{DeleteOnly}\xspace}
\newcommand{\ct}{\operatorname{count}}
\newcommand{\vsrc}{v^{\text{src}}}
\newcommand{\vtgt}{v^{\text{tgt}}}
\newcommand{\xtgt}{x^{\text{tgt}}}

\newcommand{\nl}[1]{\textit{``#1''}}

\DeclareMathOperator*{\argmin}{argmin}

\title{Delete, Retrieve, Generate: \\ A Simple Approach to Sentiment and Style Transfer}

\author{
  Juncen Li\Thanks{ Work done while the author was a visiting researcher at Stanford University.} $^1$
  \qquad Robin Jia$^2$ \qquad He He$^2$ \qquad Percy Liang$^2$ \\
  $^1$ WeChat Search Application Department, Tencent \\
  $^2$ Computer Science Department, Stanford University \\
  {\tt juncenli@tencent.com} \\ {\tt \{robinjia,hehe,pliang\}@cs.stanford.edu}
}

\date{}

\begin{document}
\maketitle
\begin{abstract}
    We consider the task of text attribute transfer:
transforming a sentence to alter
a specific attribute (e.g., sentiment)
while preserving its attribute-independent content 
(e.g., changing \nl{screen is just the right size}  to 
\nl{screen is too small}).
Our training data includes only sentences labeled with their 
attribute (e.g., positive or negative),
but not pairs of sentences that differ only in their attributes,
so we must learn to disentangle attributes 
from attribute-independent content in an unsupervised way.
Previous work using adversarial methods
has struggled to produce high-quality outputs.
In this paper, we propose simpler
methods motivated by the observation that 
text attributes are often marked by distinctive phrases (e.g., \nl{too small}).
Our strongest method extracts content words
by deleting phrases associated with 
the sentence's original attribute value, 
retrieves new phrases associated with the target attribute,
and uses a neural model
to fluently combine these into a final output. 
On human evaluation,
our best method generates grammatical and appropriate 
responses on $22\%$ more inputs
than the best previous system, averaged over
three attribute transfer datasets: 
altering sentiment of reviews on Yelp,
altering sentiment of reviews on Amazon, 
and altering image captions to be more romantic or humorous.

\end{abstract}

\section{Introduction}
\label{sec:introduction}
\begin{figure}[t]
  \begin{center}	
    \includegraphics[width=\columnwidth]{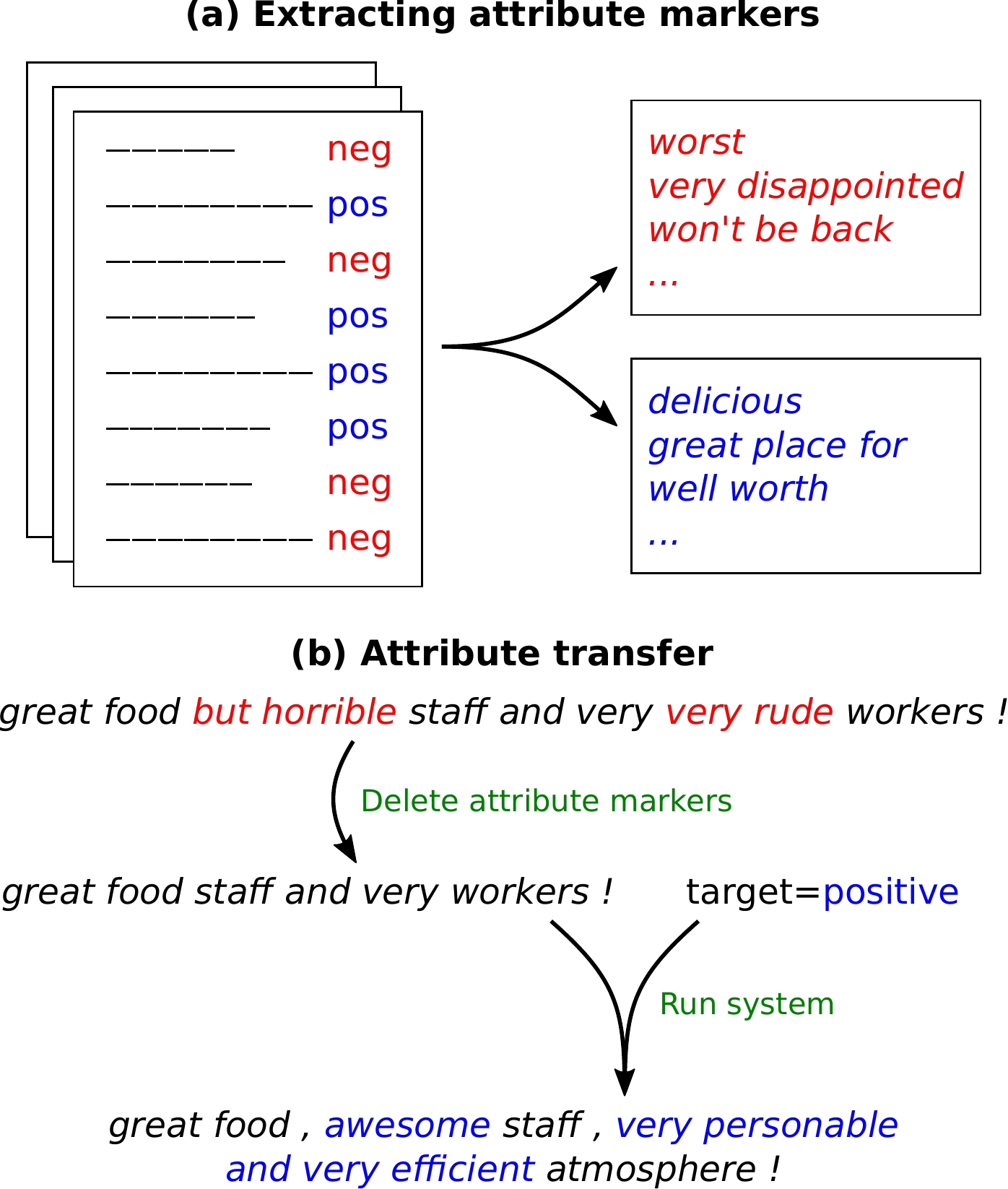}
  \end{center}
  \caption{An overview of our approach. (a) We identify
  attribute markers from an unaligned corpus.
  (b) We transfer attributes by removing
  markers of the original attribute, then 
  generating a new sentence
  conditioned on the remaining words and the target attribute.
  }
  \label{fig:overview}
\end{figure}

The success of natural language generation (NLG) systems
depends on their ability to carefully control not only the topic
of produced utterances, but also attributes such as
sentiment and style.
The desire for more sophisticated, controllable NLG has led 
to increased interest in text attribute transfer---the
task of editing a sentence to alter specific attributes,
such as style, sentiment, and tense
\citep{hu2017toward,shen2017style,fu2018style}.
In each of these cases, 
the goal is to convert a sentence with
one attribute (e.g., negative sentiment)
to one with a different attribute (e.g., positive sentiment),
while preserving all attribute-independent content\footnote{
Henceforth, we refer to attribute-independent content
as simply \emph{content}, for simplicity.}
(e.g., what properties of a restaurant are being discussed). 
Typically, aligned sentences with the same content but different
attributes are not available;
systems must learn to disentangle attributes
and content given only unaligned sentences labeled with attributes.

Previous work has attempted to use
adversarial networks \citep{shen2017style,fu2018style}
for this task, but---as we demonstrate---their outputs
tend to be low-quality, as judged by human raters.
These models are also difficult to train
\citep{salimans2016gan,arjovsky2017gan,bousmalis2017domain}.

In this work,
we propose a set of simpler, easier-to-train systems
that leverage an important observation:
attribute transfer can often be accomplished by changing a few
\emph{attribute markers}---words or phrases in the sentence
that are indicative of a particular attribute---while leaving
the rest of the sentence largely unchanged.
Figure~\ref{fig:overview} shows an example in which the sentiment of a sentence
can be altered by changing a few sentiment-specific phrases
but keeping other words fixed.

With this intuition, we first propose a simple baseline
that already outperforms prior adversarial approaches.
Consider a sentiment transfer (negative to positive) task.
First, from unaligned corpora of positive and negative sentences,
we identify attribute markers by finding phrases
that occur much more often within sentences of one attribute than the other
(e.g., \nl{worst} and \nl{very disppointed} are negative markers).
Second, given a sentence, we delete
any negative markers in it,
and regard the remaining words as its content.
Third, we retrieve a sentence with similar content from 
the positive corpus.

We further improve upon this baseline by incorporating
a neural generative model,
as shown in \reffig{overview}.
Our neural system extracts content words in the same way as our baseline,
then generates the final output with an RNN decoder
that conditions on
the extracted content and the target attribute.
This approach has significant benefits at training time,
compared to adversarial networks:
having already separated content and attribute,
we simply train our neural model to reconstruct
sentences in the training data as an auto-encoder.

We test our methods on three text attribute transfer datasets:
altering sentiment of Yelp reviews, 
altering sentiment of Amazon reviews,
and altering image captions to be more romantic or humorous.
Averaged across these three datasets, 
our simple baseline generated grammatical sentences
with appropriate content and attribute $23\%$ of the time,
according to human raters;
in contrast, the best adversarial method achieved only $12\%$.
Our best neural system in turn outperformed our baseline,
achieving an average success rate of $34\%$.
Our code and data,
including newly collected human reference outputs for the Yelp and Amazon domains,
can be found at \url{https://github.com/lijuncen/Sentiment-and-Style-Transfer}.

\section{Problem Statement}
\label{sec:problem}
We assume access to a corpus of labeled sentences 
$\mathcal{D} = \{(x_1, v_1), \dotsc, (x_m, v_m)\}$,
where $x_i$ is a sentence and $v_i \in \mathcal{V}$,
the set of possible attributes
(e.g., for sentiment,
$\mathcal{V} = \{\text{``positive''}, \text{``negative''}\}$).
We define $\mathcal{D}_v = \{x: (x, v) \in \mathcal{D}\}$,
the set of sentences in the corpus with attribute $v$.
Crucially, we do not assume access to a parallel corpus
that pairs sentences with different attributes 
and the same content.

Our goal is to learn a model that takes as input $(x,\vtgt)$
where $x$ is a sentence exhibiting source (original) attribute $\vsrc$,
and $\vtgt$ is the target attribute,
and outputs a sentence $y$ that retains the
content of $x$ while exhibiting $\vtgt$.

\section{Approach}
\label{sec:approach}
\begin{figure*}[t]
    \centering
    \includegraphics[width=\textwidth]{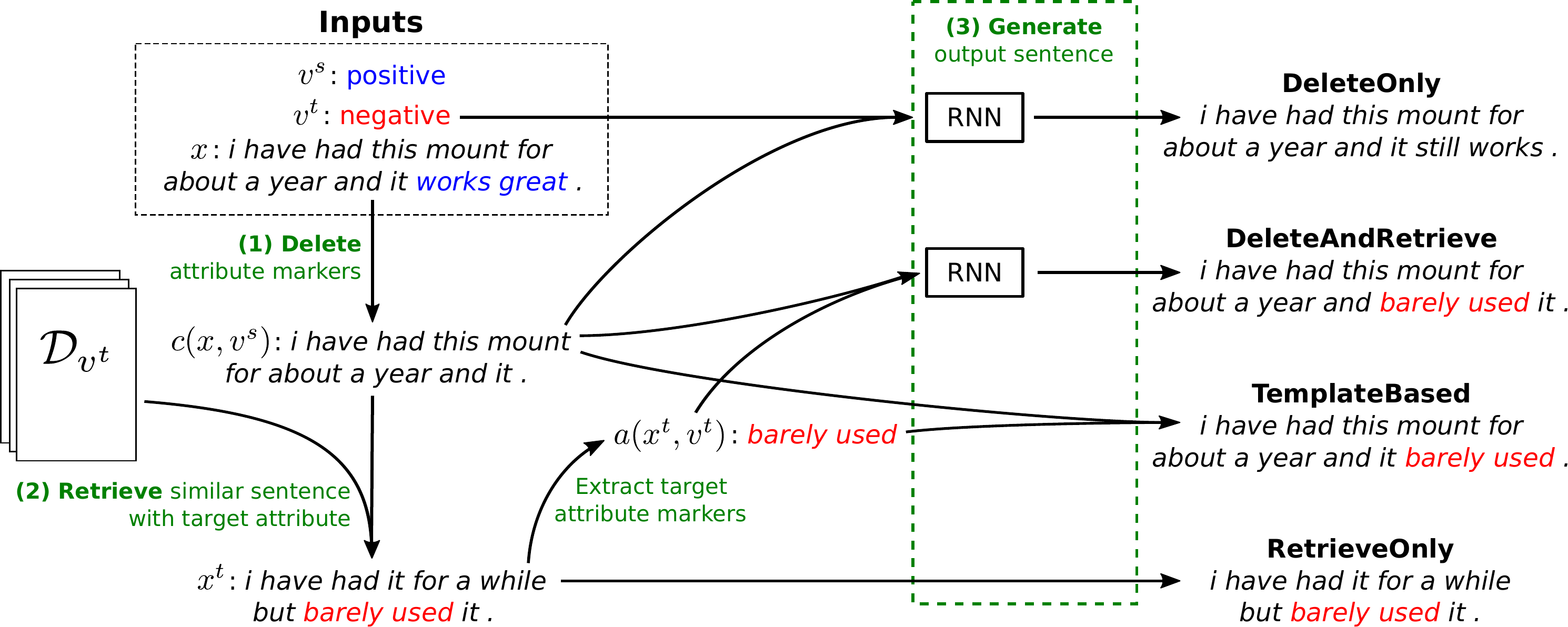}
    \caption{Our four proposed methods on the same
    sentence, taken from the \amazon dataset.
    Every method uses the same procedure (1) to separate attribute and content
    by deleting attribute markers;
    they differ in the construction of the target sentence.
    \retrieval{} directly returns the sentence retrieved in (2).
    \rulebased{} combines the content with the target attribute markers in the retrieved sentence by slot filling.
    \original{} generates the output from the content and the retrieved target attribute markers with an RNN.
    \ylabel{} generates the output from the content and the target attribute with an RNN.
    }
    \label{fig:model}
\end{figure*}
As a motivating example,
suppose we wanted to change the sentiment of \nl{The chicken was delicious.} from positive to negative.
Here the word \nl{delicious} is the only sentiment-bearing word,
so we just need to replace it 
with an appropriate negative sentiment word.
More generally, we find that the attribute is often \emph{localized} to a small fraction
of the words, an inductive bias not captured by previous work.

How do we know which negative sentiment word to insert?
The key observation is that the remaining content words provide strong cues:
given \nl{The chicken was \dots},
one can infer that a taste-related word like \nl{bland} fits,
but a word like \nl{rude} does not,
even though both have negative sentiment.
In other words, while the deleted sentiment words do contain non-sentiment
information too, this information can often be recovered using the other content words.

In the rest of this section, we describe our four systems:
two baselines (\retrieval and \rulebased)
and two neural models (\ylabel and \original).
An overview of all four systems is shown in Figure~\ref{fig:model}.
Formally, the main components of these systems are as follows:
\begin{enumerate}
  \itemsep0pt
\item
  \textbf{Delete}:
    All 4 systems use the same procedure 
    to separate the words in $x$ into 
    a set of attribute markers $a(x, \vsrc)$
    and a sequence of content words $c(x, \vsrc)$.
\item 
  \textbf{Retrieve}:
    3 of the 4 systems look through the corpus 
    and retrieve a sentence $\xtgt$
    that has the target attribute $\vtgt$ and
    %from the corpus that has the target attribute $\vtgt$
    whose content is similar to that of $x$.
\item 
  \textbf{Generate}:
    Given the content $c(x, \vsrc)$, target attribute $\vtgt$,
    and (optionally) the retrieved sentence $\xtgt$,
    each system generates $y$, either in a rule-based fashion
    or with a neural sequence-to-sequence model.
\end{enumerate}

We describe each component in detail below.

\subsection{Delete}
\label{sec:separate}
We propose a simple method to delete attribute markers ($n$-grams)
that have the most discriminative power.
Formally,
for any $v \in \mathcal{V}$, we define the \emph{salience} of an $n$-gram $u$ 
with respect to $v$ by its (smoothed)
relative frequency in $\mathcal{D}_v$:
\begin{equation}\label{eqn:style-marker}
    s(u, v) = \frac{\ct(u, \mathcal{D}_v) + \lambda}
                {\p{\sum_{v' \in \mathcal{V}, v' \ne v}
                \ct(u, \mathcal{D}_{v'})} + \lambda},
\end{equation}
where $\ct(u, \mathcal{D}_v)$ denotes the number of times an $n$-gram 
$u$ appears in $\mathcal{D}_v$,
and $\lambda$ is the smoothing parameter.
We declare $u$ to be an attribute marker for $v$
if $s(u, v)$ is larger than a specified threshold $\gamma$.
The attributed markers can be viewed as
discriminative features for a Naive Bayes classifier.

We define $a(x, \vsrc)$ to be the set of all source attribute 
markers in $x$,
and define $c(x, \vsrc)$ as the sequence of words
after deleting all markers in $a(x, \vsrc)$ from $x$.
For example, for \nl{The chicken was delicious,}
we would delete \nl{delicious} and consider \nl{The chicken was\dots}
to be the content (\reffig{model}, Step 1).

\subsection{Retrieve} 
To decide what words to insert into $c(x, \vsrc)$,
one useful strategy is to look at similar sentences with the target attribute.
For example, negative sentences that use phrases similar to
\nl{The chicken was\dots} are more likely to contain 
\nl{bland} than \nl{rude.}
Therefore, we retrieve sentences of similar content
and use target attribute markers in them for insertion.

Formally, we retrieve $\xtgt$ according to:
\begin{equation}\label{eqn:retrieve}
    \xtgt = \argmin_{x' \in \mathcal{D}_{\vtgt}} d(c(x, \vsrc), c(x', \vtgt)) ,
\end{equation}
where $d$ may be any distance metric comparing
two sequences of words.
We experiment with two options:
(i) TF-IDF weighted word overlap
and (ii) Euclidean distance using the content embeddings in Section~\ref{sec:neural} (\reffig{model}, Step 2).

\subsection{Generate}
\label{sec:neural}
Finally, we describe how each system generates $y$ (\reffig{model}, Step 3).

\textbf{\retrieval} returns the retrieved sentence $\xtgt$ verbatim.
This is guaranteed to produce a grammatical sentence with the target attribute,
but its content might not be similar to $x$.

\textbf{\rulebased}
replaces the attribute markers deleted from the source sentence $a(x, \vsrc)$ with
those of the target sentence $a(\xtgt, \vtgt)$.\footnote{
    Markers are replaced from left to right, in order.
    If there are not enough markers in $\xtgt$,
    we use an empty string.
}
This strategy relies on the assumption that if two attribute markers
appear in similar contexts \pl{that's not exactly what retrieval does, which is on the whole sentence}\hh{our retrieval only uses the content, right?},
they are roughly syntactically exchangeable.
For example, \nl{love} and \nl{don't like}
appear in similar contexts (e.g.,
\nl{i \textbf{love} this place.} and 
\nl{i \textbf{don't like} this place.}),
and exchanging them is syntactically valid.
However, this naive swapping of attribute markers
can result in ungrammatical outputs.

\textbf{\ylabel} first embeds the content
$c(x, \vsrc)$ into a vector using an RNN.
It then concatenates the final hidden state with a learned
embedding for $\vtgt$,
and feeds this into an RNN decoder to generate $y$.
The decoder attempts to produce words
indicative of the source content and target attribute, while remaining fluent.

\textbf{\original} is similar to \ylabel,
but uses the attribute markers of the retrieved sentence $\xtgt$ rather than the target attribute $\vtgt$.
Like \ylabel, it encodes $c(x, \vsrc)$ with an RNN.
It then encodes the sequence of attribute markers $a(\xtgt, \vtgt)$ with another RNN.
The RNN decoder uses the concatenation of this vector
and the content embedding to generate $y$.

\original combines the advantages of \rulebased and \ylabel.
Unlike \rulebased, \original
can pick a better place to insert the
given attribute markers,
and can add or remove function words to ensure grammaticality.
Compared to \ylabel, \original has a stronger inductive bias
towards using target attribute markers that are likely to fit in the current context.
\citet{guu2018edit} showed that retrieval strategies like ours
can help neural generative models.
Finally, \original gives us finer control over the output;
for example, we can control the degree of sentiment
by deciding whether to add \nl{good} or \nl{fantastic}
based on the retrieved sentence $\xtgt$.

\subsection{Training}
We now describe how to train \original and \ylabel.
Recall that at training time, we do not have access to 
ground truth outputs that express the target attribute.
Instead, we train \ylabel
to reconstruct the sentences in the training corpus
given their content and \emph{original} attribute value by maximizing:
\begin{equation}
\begin{aligned}
    L(\theta) = \sum_{(x, \vsrc) \in \mathcal{D}}
    \log p(x \mid c(x, \vsrc), \vsrc); \theta) .
\end{aligned}
\end{equation}

For \original, we could similarly learn an auto-encoder
that reconstructs $x$ from $c(x, \vsrc)$ and $a(x, \vsrc)$.
However, this results in a trivial solution:
because $a(x, \vsrc)$ and $c(x, \vsrc)$ were known to come from the same sentence,
the model merely learns to stitch the two sequences together without any smoothing.
Such a model would fare poorly at test time,
when we may need to alter some words to
fluently combine $a(\xtgt, \vtgt)$ with $c(x, \vsrc)$.
To address this train/test mismatch,
we adopt a denoising method similar to the 
denoising auto-encoder \cite{vincent2008denoise}.
During training, we apply some noise to $a(x, \vsrc)$
by randomly altering each attribute marker in it
independently with probability $0.1$.
Specifically, we replace an attribute marker with 
another randomly selected attribute marker of the same attribute and word-level edit distance $1$ if such a noising marker exists,
e.g., \nl{was very rude} to \nl{very rude}, which produces $a'(x,\vsrc)$.

Therefore, the training objective for \original is to maximize:
\begin{equation}\label{eqn:loss}
\begin{aligned}
    L(\theta) = 
    \sum_{(x, \vsrc) \in \mathcal{D}}
    \log p(x \mid c(x, \vsrc), a'(x, \vsrc); \theta).
\end{aligned}
\end{equation}

\section{Experiments}
\label{sec:experiments}
We evaluated our approach on three domains: 
flipping sentiment of Yelp reviews (\yelp) and Amazon reviews (\amazon),
and changing image captions to be romantic or humorous (\captions).
We compared our four systems to human references and
three previously published adversarial approaches.
As judged by human raters, 
both of our two baselines outperform all three adversarial methods.
Moreover, \original outperforms all other automatic approaches.

\subsection{Datasets}
First, we describe the three datasets we use,
which are commonly used in prior works too.
All datasets are randomly split into train, development, and test sets (Table~\ref{table:data}).

\begin{table}
  \small
  \centering
  \begin{tabular}{|c|c|r|r|r|}
    \hline
    Dataset& Attributes &Train&Dev& Test\\
    \hline
    \multirow{2}{*}{\yelp}
    &Positive&270K&2000&500 \\
    &Negative&180K&2000&500\\
    \hline
    \multirow{2}{*}{\captions} 
    &Romantic&6000&300&0 \\
    &Humorous&6000&300&0\\
    &Factual&0&0&300 \\
    \hline
    \multirow{2}{*}{\amazon}
    &Positive&277K&985&500 \\
    &Negative&278K&1015&500\\
    \hline
  \end{tabular}
  \caption{Dataset statistics.} 
  \label{table:data}
\end{table}

\paragraph{\yelp}
Each example is a sentence 
from a business review on Yelp,
and is labeled as having either positive or negative sentiment.

\paragraph{\amazon}
Similar to \yelp, each example
is a sentence from a product review on Amazon,
and is labeled as having either positive or negative sentiment~\cite{he2016amazonreview}.

\paragraph{\captions}
In the \captions dataset~\cite{gan2017style}, each example is a sentence that describes
an image, and is labeled as either factual, romantic, or humorous.
We focus on the task of converting factual sentences into 
romantic and humorous ones.
Unlike \yelp and \amazon, \captions is actually an aligned corpus---it 
contains captions for the same image in different styles.
Our systems do not use these alignments,
but we use them as gold references for evaluation.

\captions is also unique in that
we reconstruct romantic and humorous sentences during training, 
whereas at test time we are given factual captions.
We assume these factual captions carry only content,
and therefore do not look for and delete factual attribute markers;
The model essentially only inserts romantic or humorous attribute markers
as appropriate.

\subsection{Human References}
To supply human reference outputs
to which we could compare the system outputs
for \yelp and \amazon,
we hired crowdworkers on Amazon Mechanical Turk to
write gold outputs for all test sentences.
Workers were instructed 
to edit a sentence to flip its sentiment
while preserving its content.

Our delete-retrieve-generate approach relies on the prior knowledge 
that to accomplish attribute transfer,
a small number of attribute markers should be changed,
and most other words should be kept the same.
We analyzed our human reference data to understand the extent to 
which humans follow this pattern.
We measured whether humans preserved
words our system marks as content, and changed words our system marks as
attribute-related (\refsec{separate}).
We define the \emph{content word preservation rate} $S_c$
as the average fraction of words our system marks as content that were preserved by humans,
and the \emph{attribute-related word change rate} $S_a$
as the average fraction of words our system marks as attribute-related that were changed by humans:
\begin{equation}
\begin{aligned}
  S_{c} &= \frac1{|\mathcal{D}_{\text{test}}|}
  \sum_{(x, \vsrc, y^*) \in \mathcal{D}_{\text{test}}}
  \frac{|c(x, \vsrc) \cap y^*|}{|c(x, \vsrc)|}  \\
    S_{a} &= 1 - \frac1{|\mathcal{D}_{\text{test}}|}
  \sum_{(x, \vsrc, y^*) \in \mathcal{D}_{\text{test}}}
  \frac{|a(x, \vsrc) \cap y^*|}{|a(x, \vsrc)|},
\end{aligned}
\end{equation}
where $\mathcal{D}_{\text{test}}$ is the test set,
$y^*$ is the human reference sentence,
and $|\cdot|$ denotes the number of non-stopwords.
Higher values of $S_c$ and $S_a$ indicate 
that humans preserve content words and change attribute-related words,
in line with the inductive bias of our model.
$S_c$ is $0.61$, $0.71$, and $0.50$ on \yelp, \amazon, and \captions, respectively;
$S_a$ is $0.72$ on \yelp and $0.54$ on \amazon
(not applicable on \captions).

To understand why humans sometimes deviated from the inductive bias of our model,
we randomly sampled $50$ cases from \yelp
where humans changed a content word
or preserved an attribute-related word.
$70\%$ of changed content words were unimportant words
(e.g., \nl{whole} was deleted from \nl{whole experience}),
and another $18\%$ were paraphrases
(e.g., \nl{charge} became \nl{price});
the remaining $12\%$ were errors
where the system mislabeled an attribute-related word as a content word
(e.g., \nl{old} became \nl{new}).
$84\%$ of preserved attribute-related words 
did pertain to sentiment but remained fixed
due to changes in the surrounding context
(e.g., \nl{don't \textbf{like}} became \nl{\textbf{like}}, 
and \nl{below \textbf{average}} became \nl{above \textbf{average}});
the remaining $16\%$ were mistagged
by our system as being attribute-related (e.g., \nl{walked out}). 

\begin{table*}[t]
  \footnotesize
  \centering
  \begin{tabular}{|l|rrrr|rrrr|rrrr|}
    \hline
    & \multicolumn{4}{c|}{\yelp} & \multicolumn{4}{c|}{\amazon} 
    & \multicolumn{4}{c|}{\captions} \\
    & Gra & Con & Att & Suc
    & Gra & Con & Att & Suc
    & Gra & Con & Att & Suc \\
    \hline
    \crossalign
        & $2.8$ & $2.9$ & $3.5$ & $14\%$
        & $3.2$ & $2.5$ & $2.9$ & $7\%$
        & $3.9$ & $2.0$ & $3.2$ & $16\%$ \\
    \fader
        & $3.5$ & $3.7$ & $2.1$ & $9\%$
        & $3.2$ & $2.9$ & $2.8$ & $11\%$
        & $3.3$ & $2.9$ & $3.0$ & $17\%$ \\
    \multidecoder
        & $2.8$ & $3.1$ & $3.0$ & $8\%$
        & $3.0$ & $2.6$ & $2.8$ & $7\%$
        & $3.4$ & $2.8$ & $3.2$ & $18\%$ \\
    \hline
    \retrieval
        & $\bf 4.2$ & $2.7$ & $\bf 4.2$ & $25\%$
        & $3.8$ & $2.8$ & $3.1$ & $17\%$
        & $\bf 4.2$ & $2.6$ & $3.8$ & $27\%$ \\
    \rulebased
        & $3.0$ & $\bf 3.9$ & $3.9$ & $21\%$
        & $3.4$ & $3.6$ & $3.1$ & $19\%$
        & $3.3$ & $\bf 4.1$ & $3.5$ & $33\%$ \\
    \ylabel
        & $3.0$ & $3.7$ & $3.9$ & $24\%$
        & $3.7$ & $\bf 3.8$ & $3.2$ & $24\%$
        & $3.6$ & $3.5$ & $3.5$ & $32\%$ \\
    \original
        & $3.3$ & $3.7$ & $4.0$ & $\bf 29\%$
        & $\bf 3.9$ & $3.7$ & $\bf 3.4$ & $\bf 29\%$
        & $3.8$ & $3.5$ & $\bf 3.9$ & $\bf 43\%$ \\
    \hline
    Human
        & $4.6$ & $4.5$ & $4.5$ & $75\%$
        & $4.2$ & $4.0$ & $3.7$ & $44\%$
        & $4.3$ & $3.9$ & $4.0$ & $56\%$ \\
    \hline
  \end{tabular}
  \caption{Human evaluation results on all three datasets.
  We show average human ratings for grammaticality (Gra),
  content preservation (Con), and target attribute match (Att)
  on a $1$ to $5$ Likert scale, as well as overall success rate (Suc).
  On all three datasets, \original is the best overall system,
  and all four of our methods outperform previous work.
}
  \label{tab:human-eval}
\end{table*}

\subsection{Previous Methods}
We compare with three previous models, all of which use adversarial training.
\textbf{\fader}~\cite{fu2018style} learns an vector encoding of the source
sentence such that a decoder can use it to reconstruct the sentence,
but a discriminator, which tries to identify the source attribute
using this encoding, fails.  They use a basic MLP discriminator
and an LSTM decoder.
\textbf{\multidecoder}~\cite{fu2018style}
is similar to \fader,
except that it uses a different decoder for each attribute value.
\textbf{\crossalign}~\cite{shen2017style}
also encodes the source sentence into a vector,
but the discriminator looks at the hidden states of the
RNN decoder instead.  The system is trained so that the discriminator
cannot distinguish these hidden states
from those obtained by
forcing the decoder to output real sentences from the target domain;
this objective encourages the real and generated target sentences 
to look similar at a population level.

\subsection{Experimental Details}
For our methods,
we use $128$-dimensional word vectors and a
single-layer GRU with $512$ hidden units 
for both encoders and the decoder. 
We use the maxout activation function~\cite{goodfellow2013maxout}.
All parameters are initialized by sampling from a uniform distribution between $-0.1$ and $0.1$.
For optimization, we use Adadelta~\cite{matthew2012adadelta}
with a minibatch size of $256$. 

For attribute marker extraction, we consider spans up to $4$ words,
and the smoothing parameter $\lambda$ is set to $1$.
We set the attribute marker threshold $\gamma$,
which controls the precision and recall of our attribute markers,
to $15$, $5.5$ and $5$ for \yelp, \amazon, and \captions.
These values were set by manual inspection of the resulting markers
and tuning slightly on the dev set.
For retrieval, we used the TF-IDF weighted word overlap score
for \original and \rulebased, 
and the Euclidean distance of content embeddings for \retrieval.
We find the two scoring functions give similar results.

For all neural models, we do beam search with a beam size of $10$.
For \original, similar to \citet{guu2018edit},
we retrieve the top-$10$ sentences
and generate results using markers from each sentence.
We then select the output with the lowest perplexity given by
a separately-trained neural language model on the target-domain training data.

\subsection{Human Evaluation}

We hired workers on Amazon Mechanical Turk
to rate the outputs of all systems.
For each source sentence and target attribute,
the same worker was shown the output of each tested system.
Workers were asked to rate each output on three criteria on a Likert scale
from 1 to 5: grammaticality,
similarity to the target attribute, and preservation of the source content.
Finally, we consider a generated output ``successful''
if it is rated $4$ or $5$ on all three criteria.
For each dataset, we evaluated $400$ randomly sampled
examples ($200$ for each target attribute).

Table~\ref{tab:human-eval}
shows the human evaluation results.
On all three datasets, both of our baselines have a higher success rate
than the previously published models, and
\original achieves the best performance among all systems.
Additionally, we see that human raters strongly preferred the human
references to all systems, suggesting there is still
significant room for improvement on this task.

We find that a human evaluator's judgment of a sentence is largely relative to 
other sentences being evaluated together and
examples given in the instruction (different for each dataset/task). 
Therefore, evaluating all system outputs in one batch is important
and results on different datasets are not directly comparable.

\subsection{Analysis}
We analyze the strengths and weaknesses of the different systems.
\reftab{sentiment-example-outputs}
show typical outputs of each system on the \yelp and \captions dataset.

We first analyze the adversarial methods.
\crossalign{} and \multidecoder{} tend to lose the 
content of the source sentence, as seen in both 
the example outputs and the overall human ratings.
The decoder tends to generate a frequent but
only weakly related sentence with the target attribute.
On the other hand, 
\fader{} almost always generates a paraphrase of the input sentence,
implying that the encoder preserves some attribute information.
We conclude that there is a delicate balance between 
preserving the original content and dropping the original attribute,
and existing adversarial models tend to sacrifice one or the other.

Next, we analyze our baselines.
\retrieval{} scores well on grammaticality and having 
the target attribute,
since it retrieves sentences with the desired attribute
directly from the corpus.
However, it is likely to change the content when there is no perfectly aligned sentence in the target domain.
In contrast, \rulebased is good at preserving the content
because the content words are guaranteed to be kept.
However, it makes grammatical mistakes due to the unsmoothed combination of content and attribute words. 

\original{} and \ylabel{} achieve a good balance 
among grammaticality, 
preserving content, and changing the attribute.
Both have strong inductive bias on what words should be changed,
but still have the flexibility to smooth out the sentence.
The main difference is that \ylabel{} fills in
attribute words based on only the target attribute,
whereas \original{} conditions on retrieved attribute words.
When there is a diverse set of phrases to fill in---for example in \captions{}---
conditioning on retrieved attribute words helps
generate longer sentences with more specific attribute descriptions.

\begin{table*}[t]
  \newcommand{\nega}[1]{{\color{red}#1}}
  \newcommand{\posa}[1]{{\color{blue}#1}}
  \newcommand{\roma}[1]{{\color{darkgreen}#1}}
  \newcommand{\huma}[1]{{\color{purple}#1}}
  \small
  \centering
  \begin{tabular}{l|l}
    \hline
      \multicolumn{2}{c}{From \nega{negative} to \posa{positive} (\yelp) }\\
    \hline
    \source & we sit down and we got some really \nega{slow} and \nega{lazy} service .\\
    \hline
      \crossalign & we \emph{went} down and we \emph{were a \posa{good , friendly} food} .\\
      \fader   &  we sit down and we got some really \emph{\nega{slow}} and \emph{\nega{prices suck}} . \\
      \multidecoder &  we sit down and we got some really and \emph{\posa{fast} food} . \\
      \rulebased  & we sit down and we got some \emph{the service is always \posa{great}} and \emph{\posa{even better}} service .\\
      \retrieval & \emph{i} got \emph{a veggie hoagie that was \posa{massive}} and some \emph{grade a customer} service . \\
      \ylabel   & we sit down and we got some \emph{\posa{great}} and \emph{\posa{quick}} service .   \\
      \original  &we got \emph{\posa{very nice} place to} sit down and we got some service . \\
    \hline
	  \multicolumn{2}{c}{From factual to \roma{romantic} (\captions)}\\
    \hline
    \source & two dogs play by a tree .\\
    \hline
    \crossalign & \emph{a dog is running through the grass} .\\
    \fader   &  two dogs play \emph{against} a tree . \\
    \multidecoder & two dogs play by a tree . \\
    \rulebased  &two dogs play by a tree \emph{\roma{loving}} .\\
    \retrieval & two dogs \emph{are playing in a pool \roma{as best friends}} . \\
    \original  &two dogs play by a tree \emph{, \roma{enjoying the happiness of childhood}} .  \\
    \ylabel   &  two dogs \emph{\roma{in love}} play \emph{\roma{happily}} by a tree . \\
    \hline
      \multicolumn{2}{c}{From \nega{negative} to \posa{positive} (\amazon) }\\
    \hline
    \source & this is the \nega{worst} game i have come across in a long time .\\
    \hline
      \crossalign & this is the \emph{\posa{best} thing} i \emph{ve had for a few years} .\\
      \fader   &  this is the \nega{worst} game i have come across in a long time . \\
      \multidecoder &  this is the \emph{\posa{best} knife} i have \emph{no room with} a long time . \\
      \rulebased  & this is the \emph{\posa{best}} come across in a long time .\\
      \retrieval & \emph{the customer support is some of the \posa{best}} i have come across in a long time . \\
      \ylabel   & this is the \emph{\posa{best}} game i have come across in a long time .   \\
      \original  &this is the \emph{\posa{best}} game i have come across in a long time . \\
    \hline
  \end{tabular}
  \caption{Example outputs on \yelp{}, \captions{}, and \amazon.
    Additional examples for transfer from opposite directions are given in \reftab{app-example-outputs}.
    Added or changed words are in \emph{italic}.
    Attribute markers are colored.
    }
  \label{tab:sentiment-example-outputs}
\end{table*}

\subsection{Automatic Evaluation}
\label{sec:auto-eval}

\begin{table*}[t]
  \small
  \centering
  \begin{tabular}{|l|rr|rr|rr|}
    \hline
    & \multicolumn{2}{c|}{\yelp} & \multicolumn{2}{c|}{\captions} & \multicolumn{2}{c|}{\amazon}\\
    & Classifier & BLEU & Classifier & BLEU& Classifier & BLEU \\
    \hline
    \crossalign & $73.7\%$   & $3.1$ & $74.3\%$ & $0.1$ & $\textbf{74.1\%}$ & $0.4$ \\
    \fader      & $8.7\%$    & $\textbf{11.8}$ & $54.7\%$ & $6.7$ & $43.3\%$ & $10.0$\\
    \multidecoder & $47.6\%$ & $7.1$ & $68.5\%$ & $4.6$ & $68.3\%$ & $5.0$\\
    \rulebased  & $81.7\%$   & $\textbf{11.8}$ & $92.5\%$ & $\textbf{17.1}$ & $68.7\%$ & $\textbf{27.1}$\\
    \retrieval  & $\textbf{95.4\%}$ & $0.4$ & $\textbf{95.5\%}$ & $0.7$ & $\textbf{70.3\%}$ & $0.9$\\
    \ylabel     & $85.7\%$ & $7.5$ & $83.0\%$ & $9.0$ & $45.6\%$ & $24.6$\\
    \original   & $88.7\%$ & $8.4$ & $\textbf{96.8\%}$ & $7.3$ & $48.0\%$ & $22.8$\\
    \hline
  \end{tabular}
    \caption{Automatic evaluation results.
        ``Classifier'' shows the percentage of sentences labeled as the target attribute by the classifier.
        BLEU measures content similarity between the output and the human reference.
    }
  \label{tab:auto-eval}
\end{table*}
\begin{table*}[t]
  \small
  \centering
  \begin{tabular}{|c|c|c|c|}
    \hline
    \multirow{2}{*}{}&Classifier&\multicolumn{2}{c|}{BLEU} \\
    \cline{2-4}
    &Attribute&Content&Grammaticality\\
    \hline
    All data & $0.810\;(p<0.01)$&$0.876\;(p<0.01)$&$-0.127\;(p=0.58) $\\
    \hline
    \yelp& $0.991\;(p<0.01)$& $0.935\;(p<0.01)$&$0.119\;(p=0.80)$\\
    \hline
    \captions&$0.982\;(p<0.01)$&$0.991\;(p<0.01)$&$-0.631\;(p=0.13)$\\
    \hline
    \amazon&$-0.036\;(p=0.94)$&$0.857\;(p<0.01)$&$0.306\;(p=0.50)$\\
    \hline
  \end{tabular}
    \caption{
      Spearman correlation between two automatic evaluation metrics and 
      related human evaluation scores.
      While some correlations are strong, the classifier 
      exhibits poor correlation on \amazon, and BLEU only measures
      content, not grammaticality.
    }
  \label{tab:correlation}
\end{table*}

Following previous work
\citep{hu2017toward,shen2017style},
we also compute automatic evaluation metrics,
and compare these numbers to our human evaluation results.

We use an attribute classifier to assess 
whether outputs have the desired attribute
\citep{hu2017toward,shen2017style}.
We define the \emph{classifier score} as the
fraction of outputs classified as having the target attribute.
For each dataset, we train an attribute classifier on
the same training data.
Specifically, we encode the sentence into a vector by a bidirectional LSTM
with an average pooling layer over the outputs,
and train the classifier by minimizing the logistic loss.

We also compute BLEU between the output and the human references,
similar to \citet{gan2017style}.
A high BLEU score primarily indicates that the system 
can correctly preserve content
by retaining the same words from the source sentence as the reference.
One might also hope that it has some correlation with fluency,
though we expect this correlation to be much weaker.

\reftab{auto-eval}
shows the classifier and BLEU scores.
In \reftab{correlation},
we compute the system-level correlation between classifier score
and human judgments of attribute transfer,
and between BLEU and human judgments of content preservation and grammaticality.
We also plot scores given by the automatic metrics and humans in \reffig{correlation}.
While the scores are sometimes well-correlated,
the results vary significantly between datasets;
on \amazon, there is no correlation between the 
classifier score and the human evaluation.
Manual inspection shows that on \amazon,
some product genres are associated with either mostly positive
or mostly negative reviews.
However, our systems produce, for example, negative reviews
about products that are mostly discussed positively in the training set.
Therefore, the classifier often gives unreliable predictions
on system outputs.
As expected, BLEU does not correlate well with human grammaticality ratings.
The lack of automatic fluency evaluation artificially favors
systems like \rulebased, which make more grammatical mistakes.
We conclude that while these automatic evaluation 
methods are useful for model development,
they cannot replace human evaluation.

\subsection{Trading off Content versus Attribute}
One advantage of our methods is that we can control
the trade-off between matching the target attribute
and preserving the source content.
To achieve different points along this trade-off curve,
we simply vary the threshold $\gamma$ (\refsec{separate}) \emph{at test time}
to control how many attribute markers we delete from the source sentence.
In contrast, other methods~\cite{shen2017style,fu2018style}
would require retraining the model with different hyperparameters
to achieve this effect. 

\begin{figure}[t]\label{tradeoff}
  \vspace{-0.10in}
  \begin{center}	
    \includegraphics[width=3.0in]{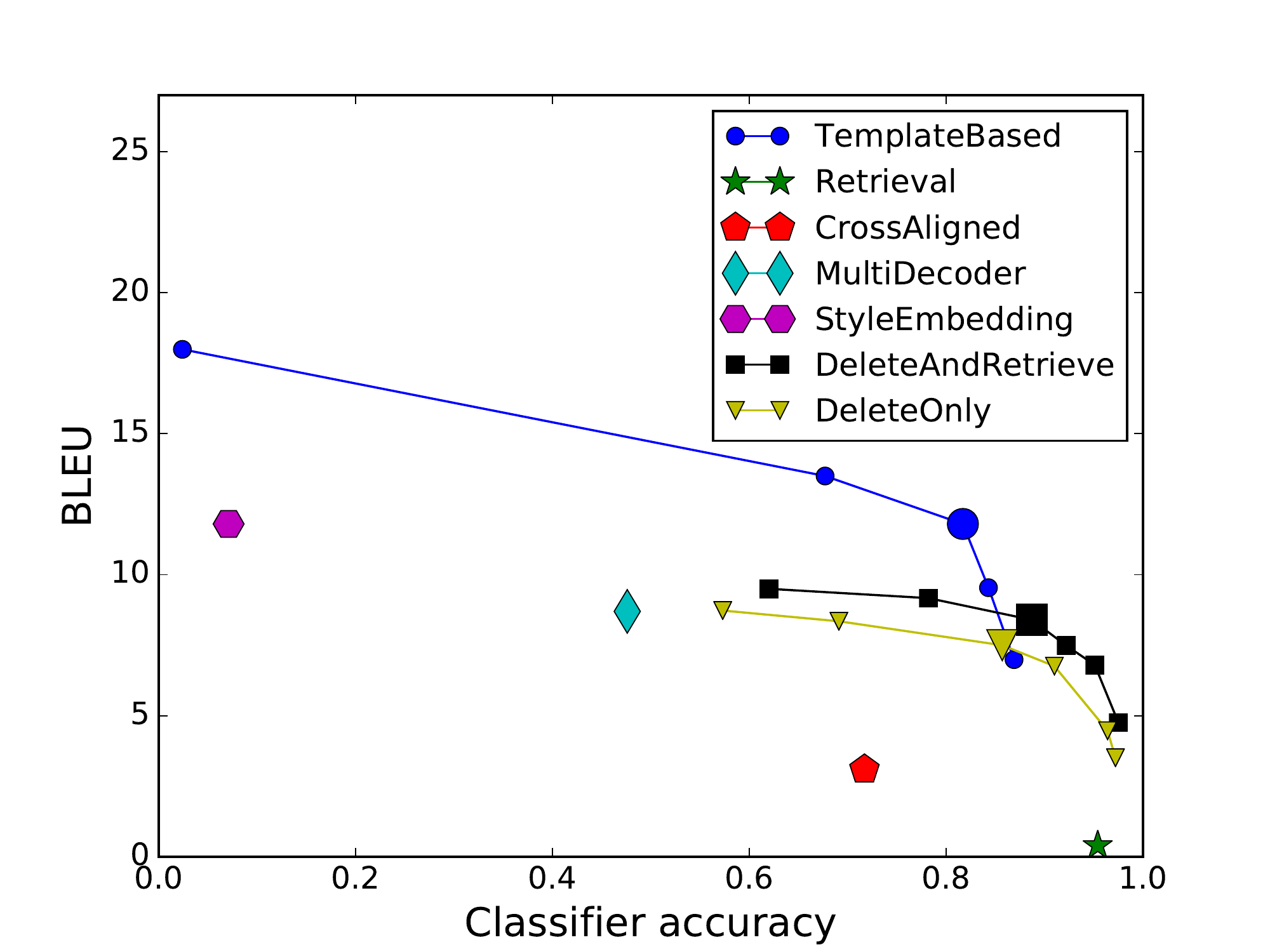}
  \end{center}
    \caption{Trade-off curves between
    matching the target attribute (measured by classifier scores)
    and preserving the content (measured by BLEU).
    Bigger points on the curve correspond to settings used
    for both training and our official evaluation.
    }
  \label{fig:tradeoff}
\end{figure}

\reffig{tradeoff} shows this trade-off curve for \original, \ylabel, and \rulebased on \yelp,
where target attribute match is measured by the classifier score
and content preservation is measured by BLEU.\footnote{
    \retrieval is less affected by what content words are preserved,
    especially when no good output sentence exists in the target corpus.
    Therefore, we found that it did not exhibit a clear content-attribute trade-off.
}
We see a clear trade-off between changing the attribute and retaining the content.

\section{Related Work and Discussion}
\label{sec:discussion}
Our work is closely related to the recent body of work on 
text attribute transfer with \emph{unaligned} data,
where the key challenge to disentangle attribute and content in an unsupervised way.
Most existing work~\cite{shen2017style,zhao2018regularized,fu2018style,igor2017attribute} uses adversarial training to separate attribute and content:
the content encoder aims to fool the attribute discriminator by removing attribute information from the content embedding.
However, we find that empirically it is often easy to fool the discriminator without actually removing the attribute information.
Therefore, we explicitly separate attribute and content
by taking advantage of the prior knowledge that
the attribute is localized to parts of the sentence.

To address the problem of unaligned data,
\citet{hu2017toward} relies on an attribute classifier to guide the generator to produce sentences with a desired attribute (e.g. sentiment, tense)
in the Variational Autoencoder (VAE) framework.
Similarly, \citet{zhao2018regularized} used a regularized autoencoder in the adversarial training framework;
however, they also find that these models require extensive hyperparameter tuning and the content tends to be changed during the transfer.
\citet{shen2017style} used a discriminator to align target sentences and sentences transfered to the target domain from the source domain.
More recently, unsupervised machine translation models~\cite{artetxe2017nmt,lample2017unsupervised}
used a cycle loss similar to~\citet{zhu2017cycle}
to ensure that the content is preserved during the transformation.
These methods often rely on bilinguial word vectors to provide word-for-word translations,
which are then finetune by back-translation.
Thus they can be used to further improve our results. 

Our method of detecting attribute markers is reminiscent
of Naive Bayes, which is a strong baseline
for tasks like sentiment classification \cite{wang2012baselines}.
Deleting these attribute markers can be viewed
as attacking a Naive Bayes classifier
by deleting the most informative features \cite{globerson2006nightmare},
similarly to how adversarial methods are trained
to fool an attribute classifier.
One difference is that our classifier is fixed,
not jointly trained with the model.

To conclude, we have described a simple method for text attribute transfer
that outperforms previous models based on adversarial training.
The main leverage comes from the inductive bias
that attributes are usually manifested in localized discriminative phrases.
While many prior works on linguistic style analysis confirm our observation that
attributes often manifest in idiosyncratic phrases~\cite{recasens2013bias,schwartz2017roc,newman2003lying},
we recognize the fact that in some problems (e.g., \citet{pavlick2017style}),
content and attribute cannot be so cleanly separated along phrase boundaries.
Looking forward,
a fruitful direction is to develop a notion of attributes more general than $n$-grams,
but with more inductive bias than arbitrary latent vectors.

\paragraph{Reproducibility.} All code, data, and experiments for this
paper are available on the CodaLab platform at
\url{https://worksheets.codalab.org/worksheets/0xe3eb416773ed4883bb737662b31b4948/}.

\paragraph{Acknowledgements.}
This work is supported by the DARPA Communicating with Computers (CwC) program under ARO prime contract no. W911NF- 15-1-0462.
J.L. is supported by Tencent.
R.J. is supported by an NSF Graduate Research
Fellowship under Grant No. DGE-114747.

\bibliographystyle{acl_natbib}
\bibliography{all}

\clearpage
\appendix

\begin{table*}[t]
  \newcommand{\nega}[1]{{\color{red}#1}}
  \newcommand{\posa}[1]{{\color{blue}#1}}
  \newcommand{\roma}[1]{{\color{darkgreen}#1}}
  \newcommand{\huma}[1]{{\color{purple}#1}}
  \small
  \centering
  \begin{tabular}{l|l}
    \hline
\multicolumn{2}{c}{From \posa{positive} to \nega{negative} (\yelp)}\\
    \hline
    \source & my husband got a ruben sandwich , \posa{he loved it} .\\
    \hline
      \crossalign & my husband got a \emph{appetizer} sandwich , \emph{she was it \nega{wrong}} .\\
      \fader   &  my husband got a \emph{\posa{greatest}} sandwich , \posa{he loved it} . \\
      \multidecoder & my husband got a \emph{beginning house with however i \nega{ignored}} . \\
      \rulebased  &my husband got a ruben sandwich , \emph{\nega{don't care}}\\
      \retrieval & i got the \emph{club} sandwich and \emph{my mom got the chicken salad sandwich} . \\
      \ylabel   & my husband got a ruben sandwich , \emph{\nega{and it was horrible !}}   \\
      \original  &my husband got a ruben sandwich , \emph{\nega{it was too dry}} . \\
    \hline
    \multicolumn{2}{c}{From factual to \huma{humorous} (\captions)}\\
    \hline
    \source & a black and white dog is running through shallow water .\\
    \crossalign & \emph{two dogs are playing on a field to win the} water .\\
    \fader   &  a black and white dog is running through shallow water . \\
    \multidecoder & a black and white dog is running through \emph{grassy} water . \\
    \rulebased  &a black and white dog is running through shallow water \emph{looking for} .\\
    \retrieval & a black and white dog is \emph{slowly} running through \emph{a snowy field} . \\
    \original  &a black and white dog is running through shallow water \emph{\huma{to search for bones}} . \\
    \ylabel   & a black and white dog is running through shallow water \emph{\huma{like a fish}} .   \\
    \hline
      \multicolumn{2}{c}{From \posa{positive} to \nega{negative} (\amazon)}\\
    \hline
    \source & i would \posa{definitely recommend} this for a cute case .\\
    \hline
      \crossalign & i would \emph{\nega{not recommend}} this for a \emph{long time} .\\
      \fader   &  i would \posa{definitely recommend} this for a cute case . \\
      \multidecoder & i would \posa{definitely recommend} this for a \emph{bra does it} . \\
      \rulebased  &\emph{\nega{skip this one}} for a cute case .\\
      \retrieval & cute \emph{while it lasted .  .  . so if you want to have a \nega{NUM night stand case} ,  this is your} case .\\
      \ylabel   & i would \emph{\nega{not recommend} it} for a cute case .   \\
      \original  &i would \emph{\nega{not recommend}} this for a cute case .\\
    \hline
  \end{tabular}
  \caption{Additional example outputs on \yelp{}, \captions{}, and \amazon.
    Added or changed words are in \emph{italic}.
    Attribute markers are colored.
    }
  \label{tab:app-example-outputs}
\end{table*}

\begin{figure*}
    \centering
    \includegraphics[width=\linewidth]{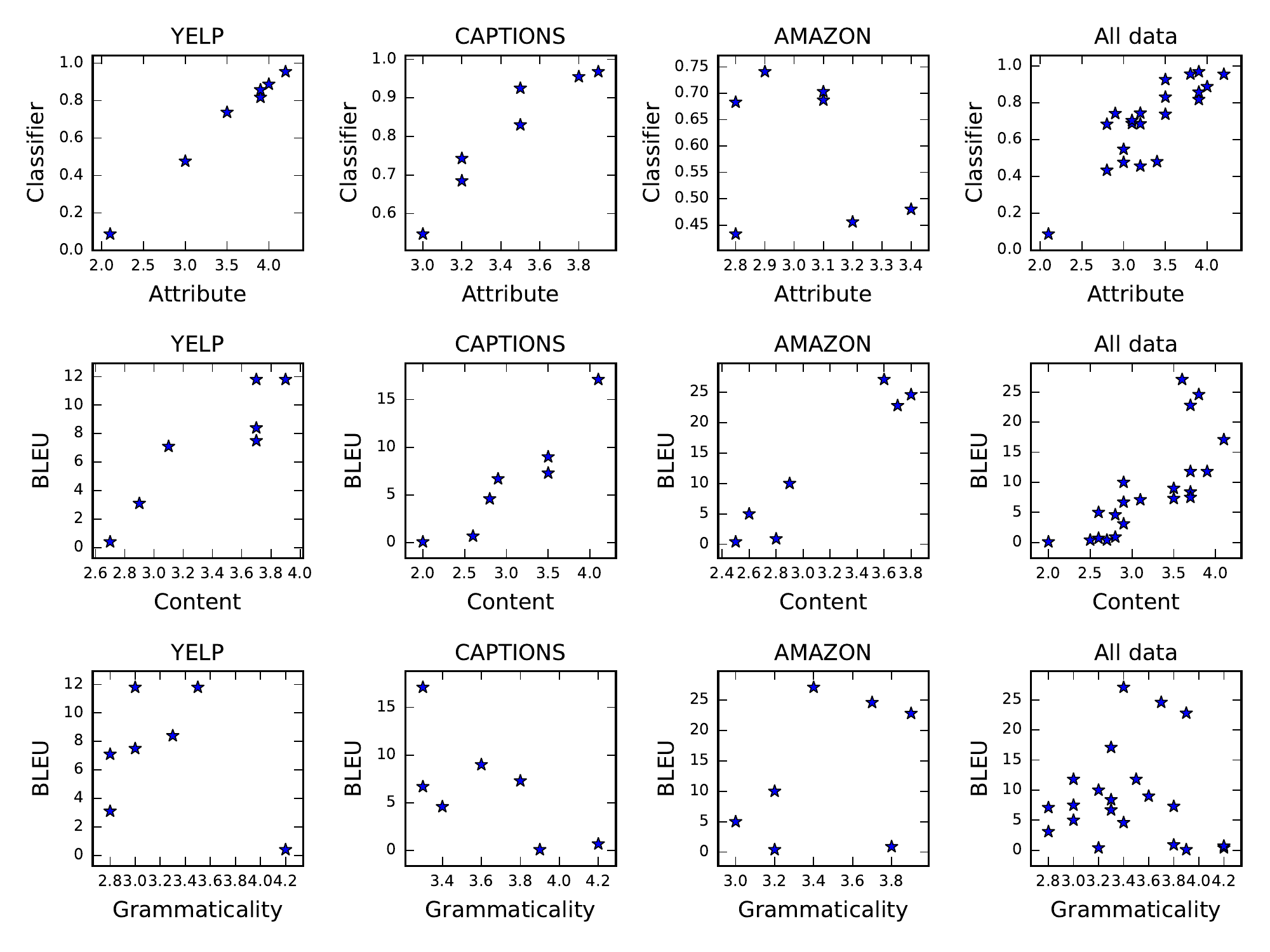}
    \caption{Scatter plots of humans scores vs. automatic metric scores on
    attribute transfer, content preservation, and grammaticality.
    The automatic metrics have some correlation with the attribute transfer
    and content preservation ratings, though results vary across datasets;
    the metrics do not measure grammaticality.
    }
    \label{fig:correlation}
\end{figure*}

\end{document}